\documentclass[10pt]{article} 

\usepackage[accepted]{rlj} 

\usepackage{amssymb}            
\usepackage{mathtools}          
\usepackage{mathrsfs}           
\usepackage{graphicx}           
\usepackage{subcaption}         
\usepackage[space]{grffile}     
\usepackage{url}                
\usepackage{lipsum}             

\usepackage{amsthm}
\usepackage{booktabs}

\title{Offline RLAIF:\\Piloting VLM Feedback for RL via SFO}

\setrunningtitle{Offline RLAIF: Piloting VLM Feedback for RL via SFO}

\author{Jacob Beck$^{1}$}

\emails{jacob\_beck@alumni.brown.edu}

\affiliations{
$^{1}$\textbf{Independent; Now at Oracle} \texttt{ | \nolinkurl{https://github.com/jacooba/OfflineRLAIF}}\\
}

\contribution{
We introduce Sub-Trajectory Filtered Behavior Cloning (SFBC), an offline RL algorithm that filters and weights sub-trajectories based on success probabilities obtained directly by querying a VLM.
To support this method, we also introduce a retrospective filtering technique that removes sub-trajectories preceding failures.
}
{
Prior work typically distills VLMs into smaller models, uses LLMs to define reward functions in code, performs direct preference optimization, or applies contrastive models to estimate goal completion heuristically.
}

\contribution{
We provide a preliminary evaluation of SFBC by comparing it to alternative VLM-based methods and standard offline RL algorithms using ground-truth rewards in a continuous control setting.
We also conduct several ablations to validate each component.
}
{
While limited to a single domain, our experiments demonstrate that there exists a standard control task where SFBC significantly outperforms more complex baselines.
}

\contribution{
We identify that preference learning on full trajectories can be uninformative in offline RL, whereas sub-trajectory supervision mitigates this issue and enables effective learning.
We corroborate this insight with empirical evidence, showing that full-trajectory preference baselines degrade both success rate and return.
}
{
To our knowledge, our work is the first to identify how RLAIF can exacerbate the \textit{stitching} problem in offline RL, and how sub-trajectory selection can resolve it.
}

\contribution{
We highlight the importance of non-Markovian feedback for RLAIF, even in Markovian environments, since VLMs must rely on visual cues over time to reason about actions they were not trained to interpret.
We empirically demonstrate that prompting for non-Markovian feedback improves performance.
}
{
To our knowledge, ours is the first work to emphasize the utility of non-Markovian feedback, even in Markovian environments, when querying VLMs directly.
}


\summary{While internet-scale image and textual data have enabled strong generalization in Vision-Language Models (VLMs), the absence of internet-scale control data has impeded the development of similar generalization in standard reinforcement learning (RL) agents.
Although VLMs are fundamentally limited in their ability to solve control tasks due to their lack of action-conditioned training data, their capacity for image understanding allows them to provide valuable feedback in RL tasks by recognizing successful outcomes.
A key challenge in Reinforcement Learning from AI Feedback (RLAIF) is determining how best to integrate VLM-derived signals into the learning process. We explore this question in the context of offline RL and introduce a class of methods called \textbf{Sub-Trajectory Filtered Optimization (SFO)}.
We identify three key insights. First, trajectory length plays a crucial role in offline RL, as full-trajectory preference learning exacerbates the \textit{stitching problem}, necessitating the use of sub-trajectories. Second, even in Markovian environments, a non-Markovian reward signal from a sequence of images is required to assess trajectory improvement, as VLMs do not interpret control actions and must rely on visual cues over time. Third, a simple yet effective approach—filtered and weighted behavior cloning—consistently outperforms more complex RLHF-based methods.
We propose \textbf{Sub-Trajectory Filtered Behavior Cloning (SFBC)}, a method that leverages VLM feedback on sub-trajectories while incorporating a 
\textit{retrospective filtering} mechanism that removes sub-trajectories preceding failures to improve robustness.
}

\begin{document}

\makeCover  
\maketitle  


\begin{abstract}
While internet-scale image and textual data have enabled strong generalization in Vision-Language Models (VLMs), the absence of internet-scale control data has impeded the development of similar generalization in standard reinforcement learning (RL) agents.
Although VLMs are fundamentally limited in their ability to solve control tasks due to their lack of action-conditioned training data, their capacity for image understanding allows them to provide valuable feedback in RL tasks by recognizing successful outcomes.
A key challenge in Reinforcement Learning from AI Feedback (RLAIF) is determining how best to integrate VLM-derived signals into the learning process. We explore this question in the context of offline RL and introduce a class of methods called \textbf{Sub-Trajectory Filtered Optimization (SFO)}.
We identify three key insights. First, trajectory length plays a crucial role in offline RL, as full-trajectory preference learning exacerbates the \textit{stitching problem}, necessitating the use of sub-trajectories. Second, even in Markovian environments, a non-Markovian reward signal from a sequence of images is required to assess trajectory improvement, as VLMs do not interpret control actions and must rely on visual cues over time. Third, a simple yet effective approach—filtered and weighted behavior cloning—consistently outperforms more complex RLHF-based methods.
We propose \textbf{Sub-Trajectory Filtered Behavior Cloning (SFBC)}, a method that leverages VLM feedback on sub-trajectories while incorporating a \textit{retrospective filtering} mechanism that removes sub-trajectories preceding failures to improve robustness.
\end{abstract}


\section{Introduction}

A long-standing goal in reinforcement learning (RL) is to develop agents capable of solving a diverse set of tasks (i.e., broad generalization) while also learning efficiently from limited experience (i.e., rapid adaptation). Despite substantial progress, RL remains highly sample-inefficient and slow to adapt to new environments. Addressing these inefficiencies requires a foundation model for RL—one that generalizes across tasks and improves learning efficiency. However, a major obstacle to achieving this is the absence of large-scale, high-quality control datasets.

Reinforcement Learning from AI Feedback (RLAIF) presents a promising approach by leveraging vision-language models (VLMs) as scalable sources of learning signals \citep{bai2022constitutional,klissarov2023motif,ma2023eureka, faldor2024omni,rocamonde2024vision, baumli2023vision,lee2024rlaif,wang2024rl}.
However, determining how best to integrate VLM-derived signals into RL remains an open challenge. A key difficulty is that VLMs struggle to discriminate between similar trajectories, making it difficult to provide precise, informative feedback. This challenge is further complicated in online RL, where agents must learn from randomly initialized policies, leading to highly ambiguous early-stage feedback.

Offline RL, in contrast, provides access to pre-collected data, which can include trajectories that are easier for VLMs to differentiate. For this reason, we focus on offline RL. However, integrating VLMs into offline RL introduces additional challenges, particularly in handling trajectory credit assignment and feedback propagation.

\begin{figure}[t]
    \centering
    \begin{subfigure}[t]{0.32\textwidth}
        \centering
        \includegraphics[width=\textwidth]{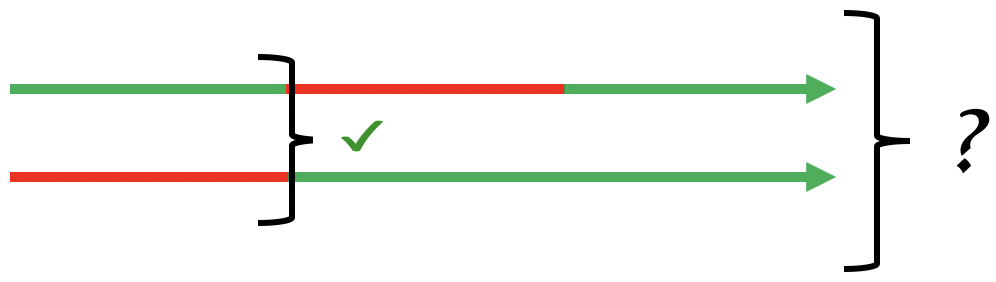}
        \caption{Comparing full trajectories can be uninformative, whereas evaluating sub-trajectories can resolve ambiguities and better capture meaningful distinctions.}
        \label{fig:sub_pref}
    \end{subfigure}
    \hfill
    \begin{subfigure}[t]{0.32\textwidth}
        \centering
        \includegraphics[width=0.3\textwidth]{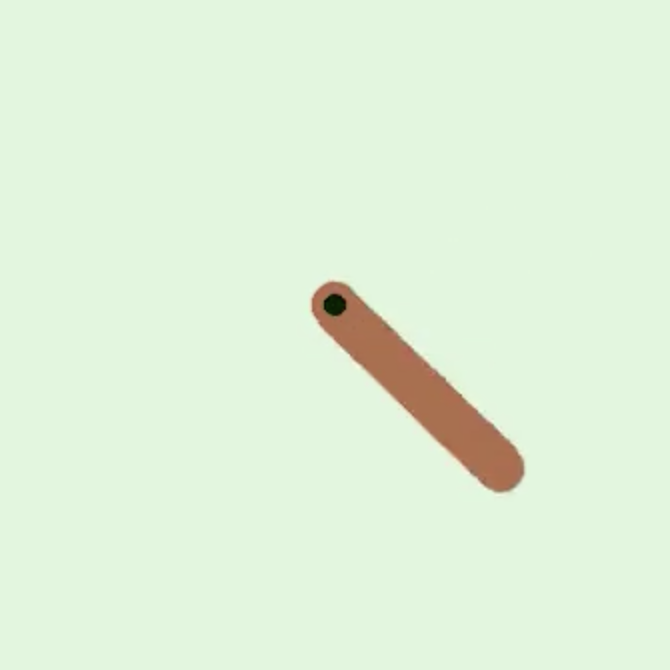}
        \includegraphics[width=0.3\textwidth]{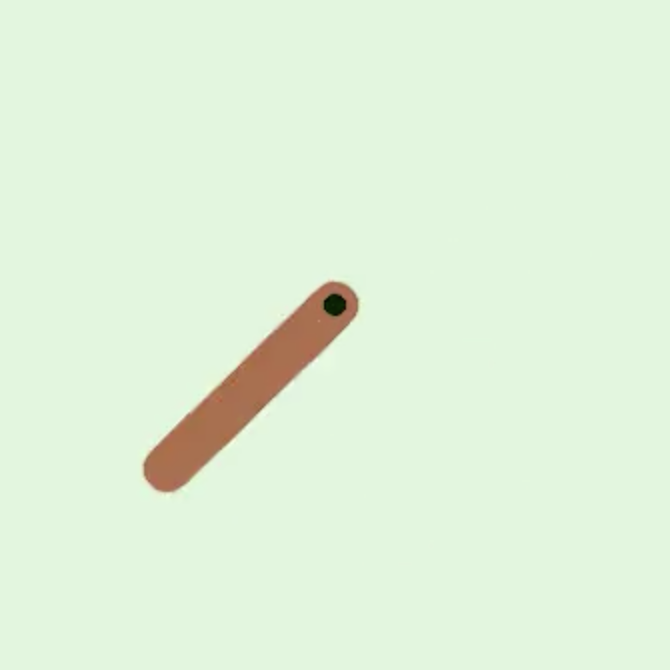}
        \includegraphics[width=0.3\textwidth]{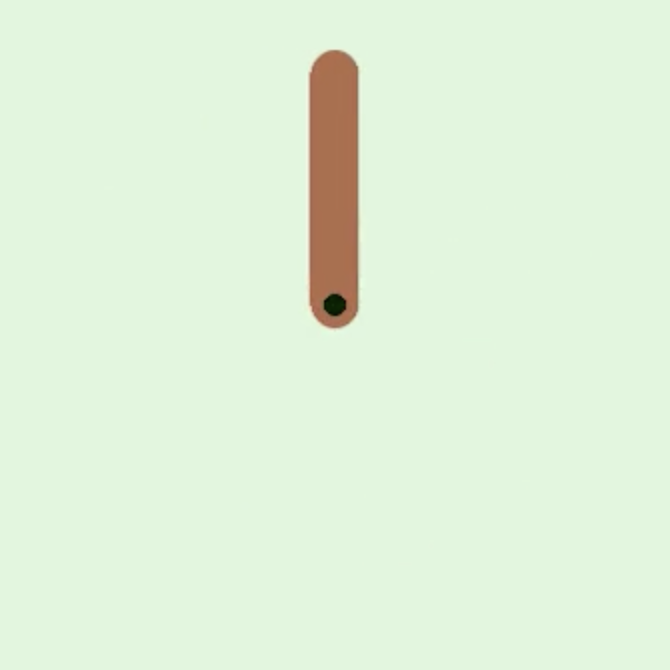}
        \caption{An example trajectory where the agent successfully swings up a pendulum, which should receive positive feedback.}
        \label{fig:good_traj}
    \end{subfigure}
    \hfill
    \begin{subfigure}[t]{0.32\textwidth}
        \centering
        \includegraphics[width=0.3\textwidth]{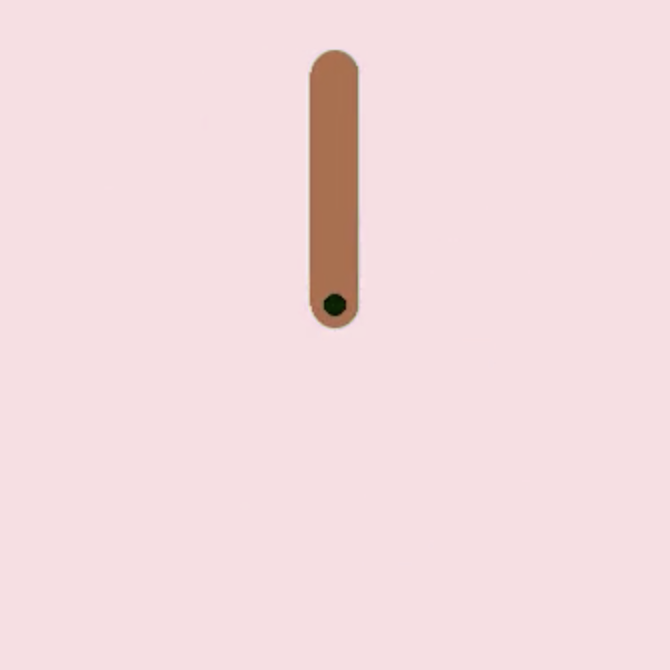}
        \includegraphics[width=0.3\textwidth]{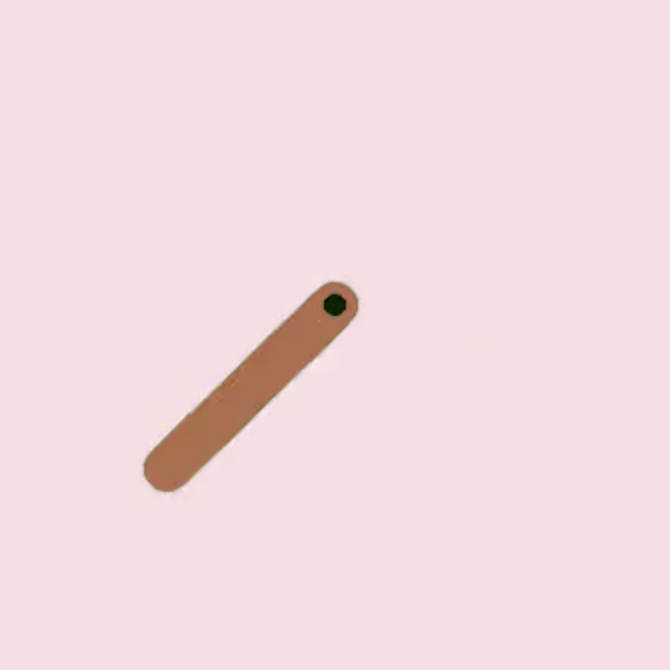}
        \includegraphics[width=0.3\textwidth]{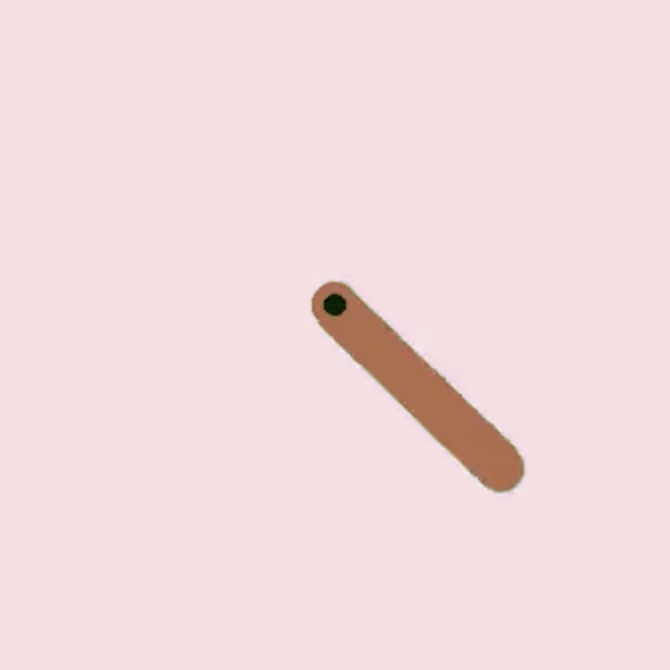}
        \caption{A failure case where Markovian feedback assigns the same reward to a bad trajectory as a good one, failing to differentiate between them.}
        \label{fig:bad_traj}
    \end{subfigure}
    \caption{Illustration of the necessity of Non-Markov sub-trajectory feedback. (a) Full-trajectory comparisons can be misleading, while sub-trajectories allow more informative distinctions. (b) A successful trajectory of a pendulum swing-up task. (c) A failure case where Markovian feedback cannot distinguish between successful and unsuccessful trajectories.}
    \label{fig:traj_comparison}
\end{figure}

To address these issues, we introduce \textbf{Sub-Trajectory Filtered Optimization (SFO)}, a class of methods that effectively incorporate VLM-derived feedback to reach new heights in offline RL. Our study identifies three key insights: 
\begin{enumerate}
    \item First, trajectory length is a crucial factor in offline RLAIF, since full-trajectory evaluation exacerbates the well-known \textit{stitching problem}. This problem necessitates the use of sub-trajectories, as depicted in Figure \ref{fig:traj_comparison}. 
    \item Second, even in Markovian environments, a non-Markovian reward model based on image sequences is required to assess trajectory improvement.
    Since VLMs are not trained on action-conditioned data, they have no basis for reasoning about how actions influence state transitions and must instead rely on visual cues over time.
    This is depicted in Figure \ref{fig:traj_comparison}.
    While discrete actions can sometimes be represented textually, many control tasks involve continuous actions (e.g., torques) that cannot be meaningfully encoded in a format suitable for VLMs.
    Furthermore, providing a reward at every time step by querying a VLM is computationally intractable for many problems without additional distillation into a smaller model \citep{wang2024rl}, and sub-sampling states to estimate reward would already abandon the Markov property.
    \item Third, despite the complexity of existing RLHF-based methods, we find that a simple yet effective approach—filtered and weighted behavior cloning—performs best in practice.
\end{enumerate}

To that end, we introduce \textbf{Sub-Trajectory Filtered Behavior Cloning (SFBC)}, which leverages VLM feedback on sub-trajectories while incorporating a retrospective filtering mechanism that removes sub-trajectories preceding failures. We empirically evaluate SFBC in a toy control domain, demonstrating its effectiveness compared to existing baselines, including naive behavior cloning and preference-based methods such as Direct Preference Optimization (DPO) \citep{rafailov2023direct}. Notably, our method outperforms AWAC \citep{nair2020awac} and TD3+BC \citep{fujimoto2021minimalist}, even when those methods use ground-truth rewards, highlighting the effectiveness of VLM-driven sub-trajectory feedback.

\section{Related Work}\label{sec:related_work}

One approach to improving RL efficiency is the manual collection of large-scale offline datasets \citep{vuong2023open}. However, this is difficult to scale due to the expense of generating diverse and high-quality trajectories on robotics in the real world. Similarly, human annotation of RL datasets is costly and difficult to scale, given the need for expert-level feedback and the challenge of obtaining consistent reward signals across tasks.

VLMs offer a promising alternative to human annotation by providing AI-generated feedback. Prior work has explored using VLMs for reward design and environment generation in code. Eureka and OMNI-EPIC \citep{ma2023eureka, faldor2024omni} have proposed methods where VLMs write programs to specify rewards and structure environments for learning. However, just as machine learning is needed to specify object categories (e.g., recognizing a dishwasher) rather than relying on hand-crafted rules, RL also requires machine-learned reward models rather than fixed rules from a VLM (e.g., to identify when a dishwasher has been unloaded). 
We firmly believe that hardcoded rules from a VLM cannot specify sufficiently complex tasks to elicit sufficiently robust reinforcement learning agents.

Recent work has also attempted to leverage AI-based reward detection to improve RLHF. Agent Q \citep{putta2024agent} uses AI-driven reward and success detection to generate preference labels for DPO. RL-VLM-F \citep{wang2024rl} distills a reward model from AI-generated labels but is an online method, and operates on Markovian states, which we find to be suboptimal. 
\citet{rocamonde2024vision} and \citet{baumli2023vision} use a CLIP model to provide Markovian feedback.
Additionally, RoboCLIP \citep{sontakke2023roboclip} encodes text and video representations using CLIP to provide feedback. While RoboCLIP-style methods offer a useful way to encode feedback, we propose an end-to-end approach using a VLM (GPT-4o in our case), which provides greater flexibility. Our study focuses on how best to utilize this feedback and how to leverage feedback in an offline context.

Complementary approaches involve automatic environment generation and world modeling \citep{bruce2024genie, valevski2024diffusion}. These efforts provide a scalable way to generate diverse training environments, but they still require specifying reward signals to learn effectively. Similarly, automated task selection \citep{wang2023voyager} and task ordering techniques such as Prioritized Level Replay (PLR) contribute to efficient learning but do not replace the need for automated high-quality feedback signals.

\section{Methods}\label{sec:methods}

Our objective is to learn an optimal policy \( \pi^* \) in an offline RL setting using vision-language model (VLM) feedback. Given a dataset \( \mathcal{D} = \{(s_t, a_t, s_{t+1}, a_{t+1},\dots)\} \) of trajectories collected by unknown behavior policies, we seek to leverage VLMs to extract meaningful rewards and preferences over trajectory segments while mitigating the challenges of stitching with offline RL and Markovian rewards with VLMs, depicted in Figure \ref{fig:traj_comparison}.

\subsection{Sub-Trajectory Filtered Behavior Cloning (SFBC)}

To incorporate VLM feedback effectively, we propose \textbf{Sub-Trajectory Filtered Behavior Cloning (SFBC)}, which refines behavior cloning by selectively weighting and filtering trajectory segments based on VLM-derived success probabilities.

\textbf{1. Sub-Trajectory Decomposition:} Each trajectory \( \tau = (s_1, a_1, s_2, a_2, \dots, s_T) \) is divided into \( n \) equal-length sub-trajectories \( \tau_i \), where:
\begin{equation}
    \tau_i = (s_{i \cdot k}, a_{i \cdot k}, s_{i \cdot k+1}, a_{i \cdot k+1}, \dots, s_{(i+1) \cdot k}),
\end{equation}
with segment length \( k = \lfloor T / n \rfloor \). Additionally, we subsample observations within each segment to reduce token usage when querying the VLM.

\textbf{2. VLM-Based Filtering:} A VLM is queried with both a Markov prompt and a non-Markov prompt. For example, in the Pendulum-v1 environment, we use:
\begin{itemize}
    \item \textbf{Markov:} ``You are watching a video of a red stick. If the black dot is at the bottom of the stick, answer `Y'. Otherwise, answer `N'.''
    \item \textbf{Non-Markov:} ``You are watching a video of a red stick. If the stick has moved between sides of the screen (left to right or right to left), answer `Y'. Otherwise, answer 'N'.''
\end{itemize}
Each prompt is followed by a sequence of images representing the sub-trajectory. The probabilities of responses (``y'') and (``yes'') (after lower-casing the response) are summed, as are the probabilities of (``n'') and (``no'').
We qualitatively find that the collective ``no’’ probability is more reliable, and both the Markov and non-Markov prompts contribute essential information. Thus, the final success probability is computed as:
\begin{align}
    P_{\text{Markov}}(\tau_i) &= 1 - P(\text{``no''} | \text{Markov Prompt}), \\
    P_{\text{Non-Markov}}(\tau_i) &= 1 - P(\text{``no''} | \text{Non-Markov Prompt}), \\
    P_{VLM}(\tau_i) &= \min\left(1, P_{\text{Markov}}(\tau_i) + P_{\text{Non-Markov}}(\tau_i) \right).
\end{align}

A sub-trajectory is retained if $ P_{VLM}(\tau_i) \geq \alpha $. To ensure robustness, we employ \textit{retrospective} filtering, where a sub-trajectory $\tau_i$ is only included if the next sub-trajectory $\tau_{i+1}$ also meets the threshold:
\begin{equation}
    \mathcal{D}_{SFBC} = \left\{ (s_t, a_t, \tau_i) \;\middle|\; \tau_i \in \mathcal{D},\; (s_t, a_t) \in \tau_i,\; P_{VLM}(\tau_i) \geq \alpha,\; P_{VLM}(\tau_{i+1}) \geq \alpha \right\}.
\end{equation}
This filtering mechanism assumes that a failed sub-trajectory likely resulted from preceding failures, ensuring that faulty transitions are not reinforced.

\textbf{3. Weighted Behavior Cloning:} 
The learned policy, $\pi_\theta$, is then trained using the following behavior cloning objective:
\begin{equation}
    \mathcal{L}_{SFBC} = -\mathbb{E}_{(s_t, a_t, \tau_i) \sim \mathcal{D}_{SFBC}} \left[ P_{VLM}(\tau_i) \log \pi_\theta(a_t | s_t) \right].
\end{equation}

\subsection{Comparison to Alternative SFO Methods}

We contrast SFBC with alternative methods leveraging sub-trajectories, including:
\begin{itemize}
    \item \textbf{VLM+TD3+BC:} Interprets $ P_{VLM}(\tau) $ as a reward and applies TD3+BC \citep{fujimoto2021minimalist}, a competitive offline RL algorithm.
    \item \textbf{S-DPO:} Applies Direct Preference Optimization (DPO) \citep{rafailov2023direct} to sub-trajectories using VLM-derived rankings.
\end{itemize}

\section{Experiments}

To evaluate the effectiveness of SFBC, we conduct experiments in the Pendulum-v1 environment, a standard continuous control benchmark. Our evaluation focuses on assessing SFBC’s ability to correctly identify and stitch together the most optimal segments from sub-optimal demonstrations while outperforming conventional offline RL baselines.
While the VLM (ChatGPT-4o) processes image-based observations to provide feedback, the learned policy itself conditions on lower-dimensional vector representations of the state, following standard RL formulations.

We compare SFBC against both standard offline RL algorithms and alternative VLM-assisted methods. To ensure robustness and statistical significance, we report success rate, mean return, and standard error across 15 seeds. Our analysis covers both overall performance comparisons and detailed ablation studies to isolate the contributions of each design choice in SFBC.

\subsection{Dataset Construction}

To evaluate the effectiveness of our proposed method in offline reinforcement learning, we construct a dataset for the Pendulum-v1 environment designed to explicitly test the ability to stitch optimal sub-trajectories together. Our dataset consists of 500 trajectories, each with a length of 600 time steps. These trajectories are generated from two distinct policies:

\begin{itemize}
    \item \textbf{Expert Policy:} A custom proportional-derivative (PD) controller designed to stabilize the pendulum in an upright position. The controller balances the pendulum when near vertical and applies a predefined acceleration-based heuristic otherwise.
    \item \textbf{Failure Policy:} A manually constructed failure policy that consistently directs the pendulum downward, ensuring suboptimal performance.
\end{itemize}

To create a dataset that challenges the ability to stitch the best segments of given trajectories, each trajectory consists of one expert policy demonstration of length 300 and one failure policy demonstration of length 300, in a random order.

\begin{table}[t]
    \centering
    \begin{tabular}{lcccc}
    \toprule
    \textbf{Method} & \textbf{Success Rate (\%)} & 
    Std. Error (\%) &
    \textbf{Mean Return} & 
    Std. Error \\
    \midrule
    BC Naive  & 33 & 12 & -4716 & 790 \\
    TD3+BC (GT) & 27 & 11 & -5131 & 814 \\
    VLM BC (Full-Trajectory) & 13 & 9 & -5234 & 578 \\
    AWAC (GT) & 0 & 0 & -7840 & 308 \\
    \midrule
    \textbf{SF-BC (Ours)} & \textbf{73} & 11 & \textbf{-1585} & 518 \\
    VLM+TD3+BC & 27 & 11 & -5013 & 649 \\
    S-DPO & 0 & 0 & -6859 & 181 \\
    \bottomrule
    \end{tabular}
    \caption{Performance comparison of SFBC against standard offline RL and VLM-assisted baselines in the Pendulum-v1 environment. SFBC achieves the highest success rate and mean return, demonstrating its effectiveness in leveraging VLM feedback for sub-trajectory optimization.}
    \label{tab:baseline}
\end{table}

\subsection{Baseline Comparisons}

We compare our proposed method, SFBC, against a set of competitive baselines, including both standard offline RL algorithms and behavior cloning methods. Specifically, we evaluate:

\begin{itemize}
    \item \textbf{BC Naive:} A standard behavior cloning baseline trained on the entire dataset, including both expert and failure trajectories, without any filtering or weighting mechanisms.
    \item \textbf{VLM BC (Full-Trajectory):} A VLM-assisted behavior cloning approach that applies success filtering at the full-trajectory level rather than sub-trajectories. Since our dataset contains a mix of expert and failure trajectories in random order, filtering at the full-trajectory level does not provide meaningful differentiation. We include this baseline primarily for didactic purposes.
    \item \textbf{TD3+BC (GT):} A strong offline RL baseline using TD3+BC \citep{fujimoto2021minimalist}, trained with access to ground-truth rewards.
    \item \textbf{AWAC (GT):} Another standard offline RL algorithm, AWAC \citep{nair2020awac}, trained on the dataset with ground-truth rewards.
\end{itemize}

The results, presented in Table \ref{tab:baseline}, demonstrate that SFBC outperforms all baselines in both mean return and success rate. Notably, SFBC achieves better performance than TD3+BC trained on ground-truth rewards, underscoring the effectiveness of leveraging VLM-derived trajectory filtering and weighting. 
This suggests that VLM-derived feedback captures non-myopic task success, effectively bypassing the need for explicit reward propagation, a known challenge in offline RL due to overestimation biases.
Additionally, SFBC surpasses S-DPO.
We hypothesize that S-DPO fails due to inherent limitations in existing vision-language models, particularly their difficulty in reasoning over multiple sequences of visual data, which is essential for accurate trajectory ranking.

Furthermore, SFBC outperforms VLM+TD3+BC, justifying our choice to treat \( P_{VLM} \) as a weight rather than a reward. 
Conceptually, the weighting scheme can be interpreted as a value function \( Q^{\pi_{\theta}} \), such that the weighted loss, $\mathcal{L}_{SFBC}$, implicitly optimizes a policy gradient. 
However, $\mathcal{L}_{SFBC}$ lacks an importance sampling correction, and \( P_{VLM} \) is computed from offline data rather than the current policy, making it a better model of \( Q^{\pi_{\text{behavior}}} \) (or \( Q^{\pi_{\text{expert}}} \) after filtering). 
SFBC also resembles AWAC \citep{nair2020awac}, except that the weighting \( P_{VLM} \) should likewise be interpreted as approximating \( e^{A^{\pi_{\text{expert}}}} \), rather than the advantage of the learned policy \( e^{A^{\pi_\theta}} \).
While SFBC does not fully correspond to a policy gradient, our results show that avoiding a direct reward interpretation—and the instability it introduces in reward propagation—is beneficial in practice.

\subsection{Ablation Studies}

We conduct ablation studies to assess the necessity of each component in SFBC. The results, presented in Table \ref{tab:ablations}, confirm that each design choice contributes significantly to performance:

\begin{itemize}
    \item \textbf{No Filtering:} Removing the filtering mechanism results in significantly worse performance, demonstrating the necessity of discarding low-confidence sub-trajectories.
    \item \textbf{Markov Prompt Only:} Using only the Markov prompt leads to degraded performance, confirming the importance of a non-Markovian understanding of trajectory improvement, even in Markovian environments, due to the inability of VLMs to process control data.
    \item \textbf{No Weighting:} Without weighting, the method lacks prioritization of high-quality sub-trajectories, leading to reduced performance.
    \item \textbf{No Retrospective Filtering:} This ablation resulted in the largest performance degradation. Without retrospective filtering, failure states remain in the dataset, potentially reinforcing poor decisions and leading to compounding errors in deployed policies. This highlights the importance of discarding preceding sub-trajectories that contribute to unsuccessful outcomes.
\end{itemize}

The ablations validate our hypothesis that all proposed modifications are necessary, further reinforcing the importance of VLM-informed filtering and weighting in offline RL.

\begin{table}[t]
    \centering
    \begin{tabular}{lcccc}
    \toprule
    \textbf{Ablation} & \textbf{Success Rate (\%)} & 
    Std. Error (\%) &
    \textbf{Mean Return} & Std. Error \\
    \midrule
    SF-BC (Ours) & \textbf{73} & 11 & \textbf{-1585} & \textbf{518} \\
    \midrule
    No Filtering  & 40 & 13 & -4164 & 883 \\
    Markov Prompt Only  & 40 & 13 & -4229 & 869 \\
    No Weighting  & 33 & 12 & -3459 & 604 \\
    No Retrospective Filtering  & 13 & 9 & -5562 & 525 \\
    \bottomrule
    \end{tabular}
    \caption{Ablation study on SFBC, evaluating the impact of key design choices. Removing filtering, non-Markovian feedback, weighting, or retrospective filtering significantly degrades performance, demonstrating the necessity of each component.}
    \label{tab:ablations}
\end{table}

\section{Conclusion}

In this work, we introduced Sub-Trajectory Filtered Behavior Cloning (SFBC), a method that leverages vision-language model (VLM) feedback for offline reinforcement learning by selectively filtering and weighting sub-trajectories. Our results demonstrate that SFBC outperforms standard offline RL methods such as TD3+BC and AWAC, as well as VLM-assisted baselines, validating the effectiveness of leveraging VLM-derived success probabilities for policy learning.

Our study provides key insights into the integration of VLM feedback with offline RL. First, sub-trajectory selection is critical, as full-trajectory preference learning exacerbates the stitching problem, making effective credit assignment impossible. Using sub-trajectories allows for improved learning stability and the ability to extract high-quality behaviors from mixed datasets. Second, non-Markovian feedback is essential, even in Markovian environments, since VLMs lack an understanding of control dynamics. Instead, they require vision-based trajectory analysis to correctly assess improvement over time. Our results confirm that non-Markovian prompts lead to significantly better performance. Finally, SFBC consistently outperforms competitive baselines, demonstrating that treating VLM-derived success probabilities as weights rather than rewards leads to superior offline policy learning. Our retrospective filtering mechanism further enhances performance by removing failure-propagating sub-trajectories. Our findings suggest that VLM feedback provides non-myopic guidance, mitigating the need for explicit reward propagation and addressing key challenges in scaling offline RL.

Most immediately, this work could be applied, at scale, to automate feedback across a sufficiently diverse set of tasks to support training a general RL foundation model from offline data.
Training the RL agent to be adaptive would inherently fall under the category of offline meta-RL \citep{beck2023survey, beck2025tutorial}.
Moreover, implementing the agent with a general-purpose sequence model, such as a transformer \citep{vaswani2017attention}, would also place it within the domain of in-context RL \citep{moeini2025survey}, while also obviating the need for complex meta-gradient computation \citep{vuorio2021no}.
Putting feedback and guidance systems on autopilot may finally provide the runway needed to arrive at a truly generalist agent for RL.

Another avenue for future research is exploring how offline feedback methods like SFBC can be extended to online RL. Since offline algorithms can be iteratively applied in an online setting, they provide a natural foundation for developing online reinforcement learning frameworks.
Finally, while we leverage a vision-language model (VLM) trained primarily for static images, future work could explore the use of models specifically trained for video understanding, which may provide more accurate and temporally consistent trajectory assessments.
As open-source and state-of-the-art VLMs continue to improve in video understanding, their ability to assess trajectory quality will become increasingly crucial for reinforcement learning applications to take off.

\appendix







\bibliography{main}
\bibliographystyle{rlj}

\beginSupplementaryMaterials

\section{Additional Experiments and Hyperparameters}

\subsection{Additional VLM+TD3+BC Experiments}

We tested a variation where filtering was applied before TD3+BC, but it did not significantly improve performance. We also tested VLM+TD3+BC with the ``Markov prompt only'' to ensure rewards for TD3 were Markovian, as well as with a discount factor of \( \gamma = 0.5 \) to encourage myopic behavior. Neither approach led to improved performance. The results are summarized in Table \ref{tab:vlm_td3_bc}.

\begin{table}[h]
    \centering
    \begin{tabular}{lccc}
    \toprule
    \textbf{Method} & \textbf{Return Mean} & \textbf{Std Err} & \textbf{Success Rate (\%)} \\
    \midrule
    VLM+TD3+BC (Standard) & -5013 & 649 & 27 \\
    VLM+TD3+BC (Filtered Before TD3+BC) & -4576 & 660 & 27 \\
     VLM+TD3+BC (Markov Prompt Only) & -4727 & 705 & 27 \\
    VLM+TD3+BC (\(\gamma = 0.5\)) & -5048 & 523 & 20 \\
    \bottomrule
    \end{tabular}
    \caption{Performance comparison of VLM+TD3+BC variants. Filtering before TD3+BC, using a Markov-only prompt, or reducing \(\gamma\) did not significantly improve performance.}
    \label{tab:vlm_td3_bc}
\end{table}

\subsection{Prompting Details}

For DPO, we used the following prompt:
\begin{quote}
    ``You are watching two videos of a red stick. The goal is to swing the stick up to gain height. It is bad to let it fall and lose momentum. Respond '1' if Video 1 is better, '2' if Video 2 is better.''
\end{quote}
Followed by:
\begin{quote}
    ``Video 1:'' $<$sub-trajectory$>$ \newline
    ``Video 2:'' $<$sub-trajectory$>$
\end{quote}

\subsection{Hyperparameters}

We used the default hyperparameters from D3RLpy for all baselines. Additionally, we tested a learning rate of \(1e^{-3}\) for AWAC and TD3+BC, as this is the default for BC, but it did not improve results. We found \( \alpha = 0.1 \) to be the most effective threshold for filtering.

For sub-trajectory processing, we used:
\begin{itemize}
    \item Sub-trajectory length: \(100\)
    \item Subsampling factor: \(20\)
\end{itemize}

Success rates, mean returns, and standard errors are reported across 15 seeds to ensure statistical robustness.


\end{document}